# Lossless Compression of Angiogram Foreground with Visual Quality Preservation of Background

Mahdi Ahmadi, Ali Emami, Mohsen Hajabdollahi, S.M.Reza Soroushmehr,
Nader Karimi, Shadrokh Samavi, Kayvan Najarian

*Abstract*— **By increasing the volume of telemedicine information, the need for medical image compression has become more important. In angiographic images, a small ratio of the entire image usually belongs to the vasculature that provides crucial information for diagnosis. Other parts of the image are diagnostically less important and can be compressed with higher compression ratio. However, the quality of those parts affect the visual perception of the image as well. Existing methods compress foreground and background of angiographic images using different techniques. In this paper we first utilize convolutional neural network to segment vessels and then represent a hierarchical block processing algorithm capable of both eliminating the background redundancies and preserving the overall visual quality of angiograms.**

## I. INTRODUCTION

Angiogram images are an important source of information for diagnosis of coronary artery diseases. With recent progress in technology, we observe the advent of distance medical information transfer. By increasing the demand for transferring medical images among hospitals, the medical image compression has gained a lot of attention. Image compression methods can be lossy or lossless. Although an extracted image in lossy methods is not identical to the original one, we get good compression ratio in return. These methods often work based on removing redundancies in frequency domain, such as DCT coefficients, wavelet domain, etc. One example of lossy image compression methods is proposed in [1] which works on wavelet domain coefficients. Another example is the method proposed in [2] which takes into account the DCT coefficients of image blocks. However, in the category of lossless methods, original images can be completely recovered. Hence, these methods are suitable for medical applications. Many of the proposed algorithms in this class of media compression, work in spatial domain by using prediction of pixels. One example of lossless image compression is presented in [3] which employs different lossless algorithms and entropy coding.

We may notice that lossless compression is not necessary for the entire image. For example, in angiography images, a large area of each frame is considered as background, while major arteries must be fully preserved for stenosis/blockage diagnosis. In order to preserve diagnostically important regions such as arteries and compress them losslessly, *diagnostically lossless compression* (DLC) methods have been proposed. In DLC methods an image is first divided to two parts, *region of interest* (ROI) that includes crucial regions in terms of diagnosis and a non-ROI (NROI) that contains background. Then, for compressing each part different methods are applied. Qi and Tyler propose a DLC method that does not necessarily preserve the ROI completely; however, the reconstructed ROI is less damaged than NROI [4]. Chaabouni *et al.* [5] present a DLC method in which ROI and Non-ROI are compressed using high and low resolution versions of incremental self-organizing map respectively [6]. Shuai *et al.* [7] employ Shearlet transform to compress ROI and fractional encoding in wavelet transform to compress NROI. In another DLC method, a tree weighing lossless compression and fractal lossy compression methods are applied on ROI and NROI respectively [8]. Compressing ROI can be performed by JPEG-LS [3] and a wavelet based lossy-compression algorithm can be employed for NROI [9]. Ström *et al.* compress ROI and NROI by using S-transform and lossy wavelet zero-tree respectively [10]. Another approach is to suppress the background by setting NROI to zero and then compress the resulting image by a lossless method [11]. In this approach, the background information, that might be informative for diagnosis, is ignored causing image appearance and visual perception degradation.

In this paper angiography images are segmented into ROI and NROI using our proposed convolutional neural network (CNN) framework [12]. Then, we propose a hierarchical approach to partition each of those regions to blocks/sub-blocks. For the ROI blocks, we reduce redundancies by eliminating specific DCT coefficients. In order to decide which coefficients can be removed, we employ the Canny edge detector and to avoid typical blockiness artifact of the transform domain compression method, we utilize a smoothing filter. This filter not only enhances visual aspects of the image, but also removes further redundancies. Afterwards, the JPEG-LS compression [3] method is applied to the resulting image. Similar to other above-mentioned methods, the proposed method is a tradeoff between compression ratio and the preservation of information.

The remainder of this paper is organized as follows: in section II, we explain our method in details. In section III,

M. Ahmadi, M. Hajabdollahi, and N. Karimi, are with the Department of Electrical and Computer Engineering, Isfahan University of Technology, Isfahan 84156-83111, Iran.

A. Emami is with the Department of Electrical and Computer Engineering, Isfahan University of Technology, Isfahan 84156-83111, Iran. He is also with the Dept. of Information Technology and Elect. Engineering, University of Queensland, Brisbane, Australia.

S.M.R. Soroushmehr is with the Department of Computational Medicine and Bioinformatics and the Michigan Center for Integrative Research in Critical Care, University of Michigan, Ann Arbor, MI, U.S.A.

S. Samavi is with the Department of Electrical and Computer Engineering, Isfahan University of Technology, Isfahan 84156-83111, Iran. He is also with the Department of Emergency Medicine, University of Michigan, Ann Arbor, MI, U.S.A.

K. Najarian is with the Department of Computational Medicine and Bioinformatics, Department of Emergency Medicine and the Michigan Center for Integrative Research in Critical Care, University of Michigan, Ann Arbor, MI, U.S.A.

experimental results are represented. We conclude the paper in section IV with a brief discussion and summary.

## II. THE PROPOSED METHOD

Overview of the proposed method for medical image compression is illustrated in Fig. 1. For image segmentation we utilize the proposed CNN framework [12]. This network gets gray-scale angiography images as input and produces an ROI probability map, which estimates the foreground/ROI regions. Using this map, we partition NROI pixels to blocks and reduce the spatial redundancies in the blocks by removing DCT coefficients. Hence, the prediction based lossless algorithms can work on the resulting image effectively, as most of the redundancies in NROI (background areas) have already been removed.

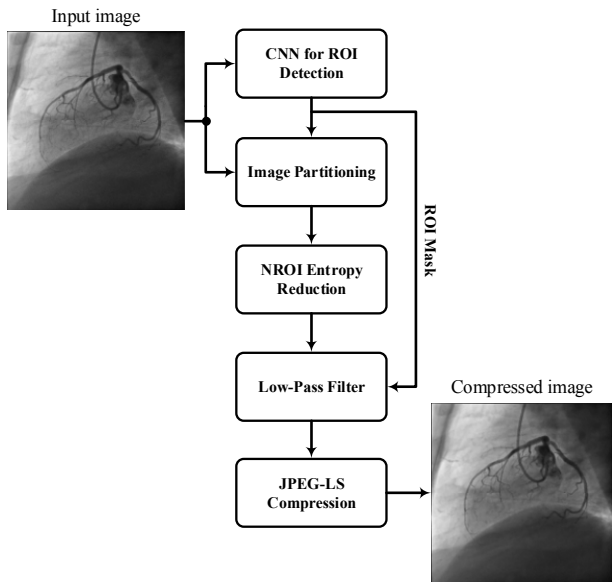

Figure 1: the overall diagram of the proposed method.

### A. CNN for ROI detection

For image segmentation we utilize our previously designed CNN presented in [12]. This network is a patch-based network that gets a gray-scale angiography image patch with the size of 33×33 and produces a probability map showing the probability of a pixel belonging to the ROI. As suggested in [12], the segmented vessels are obtained by applying a threshold on the generated probability map. Then, the largest connected component is considered as the ROI and other components as NROI.

### B. Image partitioning

In this part, we partition an image to 8 × 8 blocks, each of which consists of 4 sub-blocks. A block or sub-block is called an NROI block if all of its pixels belong to NROI and is called an ROI block if at least one of its pixels belongs to ROI. For ROI blocks we investigate their sub-blocks and check if that sub-block is an ROI sub-block or not. For the ROI blocks and sub-blocks we don't change anything. As shown in Fig. 2, this hierarchical algorithm lets us reduce entropy for more regions of NROI. In Fig. 3, the application of image partitioning for entropy reduction procedure is illustrated.

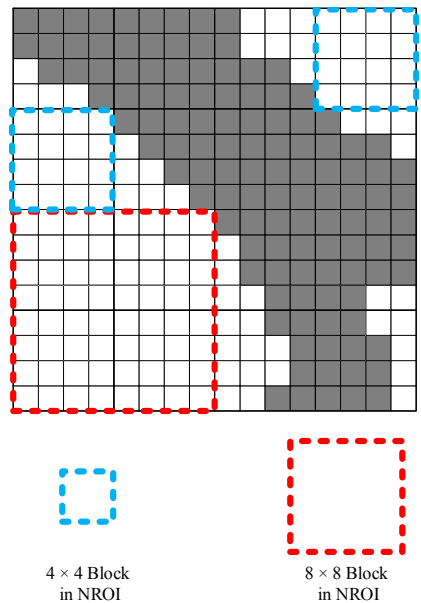

Figure 2: the hierarchical block representation of the image

### C. Non ROI entropy reduction

In order to remove redundancies, we apply the DCT coefficients elimination method. For this purpose, we replace higher frequency DCT coefficients with zeros in NROI blocks. We show in section III that this reduction operation reduces the size of output image after the final lossless compression (JPEG-LS algorithm) is accomplished. We use city block distance of a coefficient from the origin of the block to decide if the coefficient is low. For this purpose we use zigzag coefficient sweeping similar to JPEG standard [13]. Then we keep the first $\gamma$ coefficients of DCT, where $\gamma$ is calculated by equation (1).

$$\gamma = \begin{cases} \max(64, \gamma_0 + N_{Edge}) & \text{for } 8\times 8 \text{ blocks} \\ \max(16, \gamma_0 + N_{Edge}) & \text{for } 4\times 4 \text{ blocks} \end{cases} \quad (1)$$

As can be seen in (1), we assume that the blocks with more edges are more important. The parameter $\gamma_0$ in (1) is a variable that can be tuned. The term $N_{Edge}$ is the number of edge pixels in a block where edges are obtained by Canny edge detection method [14]. For this purpose we find edges throughout the whole image, prior to partitioning it into blocks and sub-blocks. If the effect of edges is to be ignored, we simply ignore the $N_{Edge}$ term. The last step in this stage of the algorithm is to tile blocks and sub-block to form the reduced entropy image.

### D. Low pass filtering and JPEG-LS Compression

After tiling of ROI and Non ROI blocks, the image may have blockiness effects. To remove this effect a smoothing procedure is applied by the Gaussian filter. In addition to suppressing the unwanted noises in the original image, the Gaussian filter results better compression after JPEG-LS. However, we do not eliminate it from the ROI pixels because it can lead to misdiagnosis of the physician. In order to have accurate and non-smoothed ROI pixels, we then replace the ROI pixels in filtered image with the original ones. Finally the JPEG-LS compression is applied to compress the NROI smoothed image.

## III. EXPERIMENTAL RESULTS

We test our framework on the angiogram dataset used in [12], which contains 44 images of size $512 \times 512$ pixels.

The proposed method in this paper is based on a tradeoff between PSNR (Peak Signal to Noise Ratio) and compression ratio. Let us assume that the JPEG-LS size of the original image is called $N_O$ and the JPEG-LS size of the image that we reduced its background entropy is $N_B$. Then the compression ratio ($CR$) is defined as $CR = \frac{N_O}{N_B}$.

On one end, we may preserve all the image details to maximize PSNR or we can eliminate the background information to increase the compression ratio. We evaluate the system objectives, compression ratio and PSNR, by averaging the calculated metrics over all dataset images. The PSNR metric is calculated by the following equation:

$$PSNR = 10\, log_{10}\left(\frac{MAX_{orig}^2}{MSE}\right) \qquad (2)$$

where $MAX_{orig}$ is the maximum intensity value of the original image (255 for 8 bit images) and MSE is the mean square error of the ROI pixels calculated by (3)

$$MSE = \frac{1}{N_G} \sum_{(i,j)\in ROI_G} \left(I_{orig}(i,j) - I_{proc}(i,j)\right)^2 \qquad (3)$$

where $I_{orig}$ and $I_{proc}$ are original and processed images respectively and $N_G$ is the number of pixels on the ROI's ground-truth ($ROI_G$). The PSNR is calculated on the ROI's ground-truth. Our obtained ROI may be different from the ground truth. Fig.4 illustrates the tradeoff between compression ratio and PSNR, for different values of $\gamma_0$.

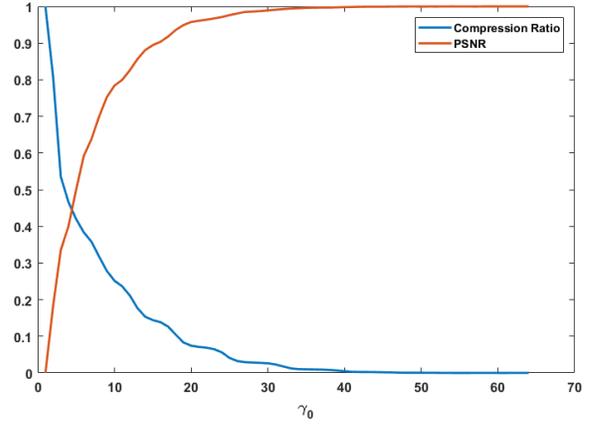

Figure 4: Tradeoff between normalized compression ratio and normalized PSNR.

In order to evaluate the impact of taking edges into account in (1), we first perform two simulations, one of them with using the Canny edge detector and the other one without that. The results for the parameter $\gamma_0$ in 8×8 blocks are shown in Fig.5 where each circle denotes a choice for $\gamma_0$. The top-left circle of each color is for $\gamma_0 = 1$ and the bottom-right circle is for $\gamma_0 = 64$. As demonstrated in Fig. 5, considering Canny edges would significantly improve the performance of our proposed method. For example, for a fixed compression ratio the PSNR of the blue curve is more than one of the red curve.

To have a good criterion for choosing $\gamma_0$ the graph of Fig.6 is plotted. We add the normalized PSNR and compression ratio values of Fig. 5 to generate the graph of Fig. 6. A suitable range of values for $\gamma_0$ should correspond to very high values on the graph of Fig. 6. According to Fig.6, a good range for selecting $\gamma_0$ is from 8 to 25. From our experiments we keep all 16 DCT coefficient in 4×4 blocks. This lets us have fewer distortions in the ROI boundaries. Hence, the only process on the 4×4 blocks is the Gaussian filter. The other tunable parameter is $\sigma$ (variance) of the Gaussian filter. Higher values of $\sigma$ lead to higher compression ratios; however it has the disadvantage of reduction of the PSNR and visual quality of the image. On the other hand, by lowering the value of $\sigma$ beyond a point, the blockiness effects start to appear. The threshold value on selecting $\sigma$ is based on $\gamma$ and also the amount of blockiness is

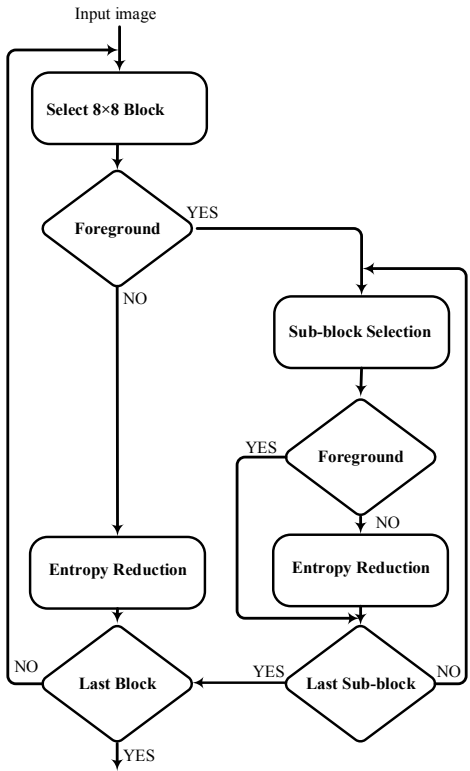

Figure 3: Proposed hierarchical block selection and processing.

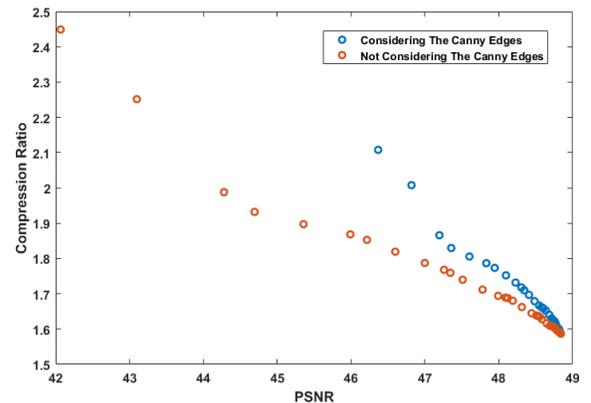

Figure 5: Effects of including the $N_{Edge}$ in (1).

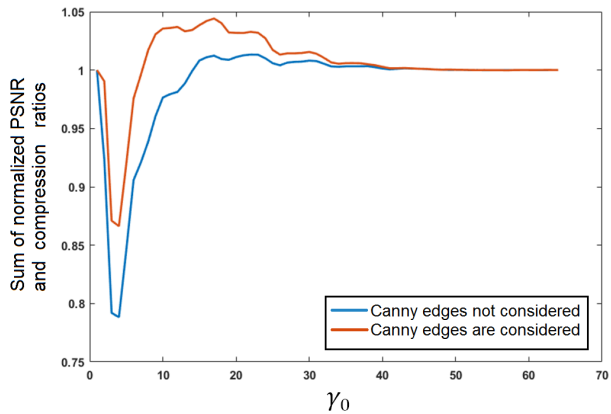
Figure 6: Multi-objective diagram of the proposed method.

subjective i.e. we do not have a metric for it. According to the visual quality of images we can say a good selection of $\sigma$ parameter would be 1.

For the Canny edge detection, our choice in selection of the threshold value was 0.15. As the threshold value is lowered, the number of detected edges is increased. Consequently, a higher PSNR is obtained but this would lower the compression ratio. The hierarchical procedure discussed in this paper can be continued to have smaller sub-blocks, however we did not observe significant benefits in it. Fig.7 shows a sample image and its compressed versions by our method and standard JPEG [13]. As seen in Fig.7, our method has less distortion on ROI.

## IV. CONCLUSION

In this paper we proposed a diagnostically lossless medical image compression method. We segmented the image to ROI and NROI utilizing convolutional neural networks. Then, by applying a hierarchical block processing algorithm on the NROI region, we reduced its redundancy. The processed NROI was then merged with the unchanged ROI and the resulting image was passed to the JPEG-LS coder. Our experimental results show that the compressed results have less size than non-processed image after performing JPEG-LS. The method however affects the image quality especially in the NROI part.

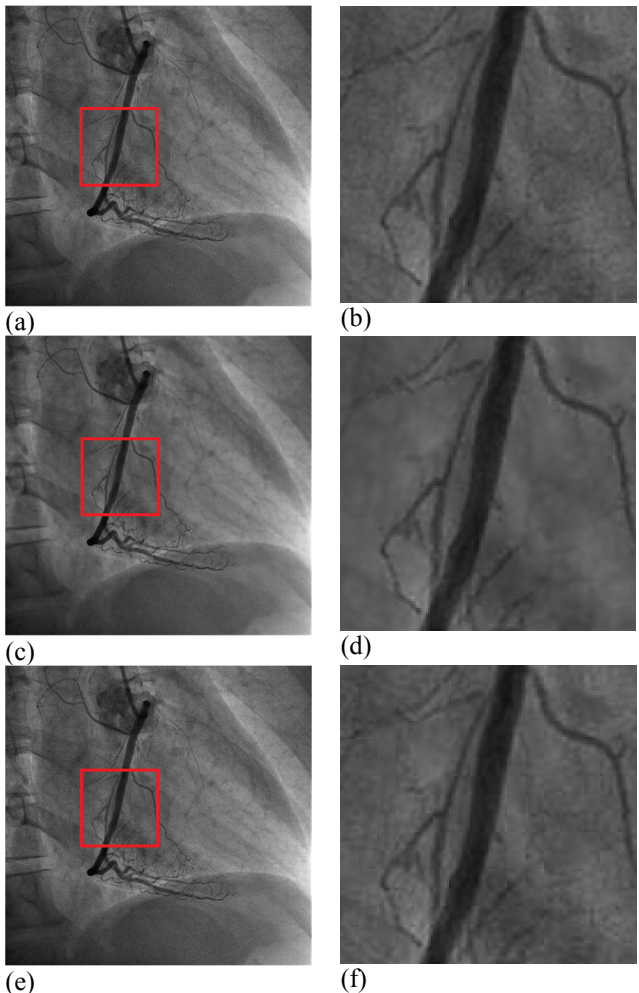
Figure 7: Left column from top to bottom: Original image, Result after performing our compression method, Result after performing standard JPEG (Q=60). Right column: Zoomed versions of left images.